# labelCloud: A Lightweight Domain-Independent Labeling Tool for 3D Object Detection in Point Clouds

Christoph Sager, christoph.sager@gmail.com, Technische Universität Dresden
Patrick Zschech, patrick.zschech@fau.de, Friedrich-Alexander-Universität Erlangen-Nürnberg
Niklas Kühl, niklas.kuehl@kit.edu, Karlsruhe Institute of Technology

**Introduction**

Within the past decade, the rise of applications based on artificial intelligence (AI) in general and machine learning (ML) in specific has led to many significant contributions within different domains. The applications range from robotics over medical diagnoses up to autonomous driving. However, nearly all applications rely on trained data. In case this data consists of 3D images, it is of utmost importance that the labeling is as accurate as possible to ensure high-quality outcomes of the ML models. Labeling in the 3D space is mostly manual work performed by expert workers, where they draw 3D bounding boxes around target objects the ML model should later automatically identify, e.g., pedestrians for autonomous driving or cancer cells within radiography.

While a small range of recent 3D labeling tools exist, they all share three major shortcomings: (i) they are specified for autonomous driving applications, (ii) they lack convenience and comfort functions, and (iii) they have high dependencies and little flexibility in data format. Therefore, we propose a novel labeling tool for 3D object detection in point clouds to address these shortcomings.

**Fundamentals and Previous Approaches**

The recent progress in autonomous driving was enabled by two major developments: advances in ML and improved sensors. Especially 3D sensors like light detection and ranging (LiDAR) and depth cameras have become more precise and affordable over the years. These devices create 3D reconstructions of their environment by either measuring the time of flight from laser rays (i.e., LiDAR) or comparing the images from parallel cameras (i.e., stereovision). With this technology, it has become easier to tell the distance to and size of surrounding objects, as the parameters do not have to be guessed from two-dimensional images anymore. Moreover, the ubiquity of 3D sensors has made this technology interesting for multiple domains (e.g., robotics, medicine, virtual reality).

With the integration of LiDAR into latest smartphones, the technology is now spreading into millions of devices, which has sparked interest in the automatic processing and understanding of the data generated. Most 3D sensors output point clouds – unordered sets of points in Euclidean space. The unordered nature of this data type and the absence of any fixed grid (like in 2D images) has made it difficult to simply lift successful solutions from 2D computer vision into the 3D space. However, a new architecture called PointNet [4] has led to a breakthrough and enabled multiple solutions that can detect objects directly inside the point cloud data [2]. 3D object detection methods can automatically identify and locate objects with their class, position, dimension, and sometimes even rotation. Still, all existing approaches are based on supervised ML, where an artificial neural network gets trained for a specific problem by providing many examples with known outcomes. Consequently, researchers and practitioners first have to label large amounts of training data to create accurate ML models.

Existing non-commercial point cloud labeling tools focus exclusively on the domain of autonomous driving [1][5][7]. Thus, they only support data formats and object types that commonly occur in this specific domain. Furthermore, as vehicles tend to be parallel to the floor, most labeling tools consider only the object rotation around the vertical axis. In doing so, they reduce the dimensionality to a 2D selection from a top-down perspective and infer the bounding box height indirectly. Direct labeling on the other hand lets the user directly draw a 3D bounding box inside the point cloud. We developed labelCloud, a lightweight and domain-independent labeling tool for annotating rotated bounding boxes in 3D point clouds. labelCloud supports LiDAR sensors and depth cameras (with seven input formats), multiple label formats ready for use in existing ML frameworks, and the rotation of bounding boxes around all three axes for 6D pose estimation. Table 1 compares our solution with existing approaches.

|  |  | 3D Bat [7] | LATTE [5] | SAnE [1] | labelCloud |
|---|---|---|---|---|---|
| Input | *.bin | ✗ | ✓ | ✓ | ✓ |
|  | *.ply | ✗ | ✗ | ✗ | ✓ |
|  | *.pcd | ✓ | ✗ | ✗ | ✓ |
|  | *.xyz | ✗ | ✗ | ✗ | ✓ |
|  | LiDAR sensors | ✓ | ✓ | ✓ | ✓ |
|  | 3D cameras | ✓ | ✗ | ✗ | ✓ |
|  | Colored | ✓ | ✗ | ✗ | ✓ |
| Labeling | Direct | ✓ | ✗ | ✗ | ✓ |
|  | x-rotation | ✗ | ✗ | ✗ | ✓ |
|  | y-rotation | ✗ | ✗ | ✗ | ✓ |
|  | z-rotation | ✓ | ✓ | ✓ | ✓ |
|  | Orientation | ✓ | (✓) | (✓) | ✓ |
| Setup | Dependencies | 8 | 37 | 47 | 4 |
|  | No proprietary libraries | ✗ | ✓ | ✓ | ✓ |
|  | No preparation needed | ✗ | ✗ | ✓ | ✓ |
|  | No trained models needed | ✗ | ✗ | ✗ | ✓ |
| Support | Object tracking | ✓ | ✓ | ✓ | ✗ |
|  | Point cloud alignment | ✗ | ✗ | ✗ | ✓ |
|  | Code available | GitHub | GitHub | GitHub | GitHub |
|  | Fulfilled criteria | 8/17 | 6/17 | 7/17 | 16/17 |

Tab. 1: Comparison of existing point cloud labeling tools for a domain-independent usage.

**Main Idea**

The project was initiated due to the lack of suitable labeling tools for annotating colored point clouds as commonly generated by 3D cameras (like the Intel RealSense series). Researchers wanting to tap into that data faced the problem that existing software was either very complex to set up or incompatible with the desired data format (like *.ply or *.pcd). Consequently, we developed a solution that would let users quickly generate annotated training data from their depth sensors to accelerate the research in 3D vision—independent of the respective domain. First, we conducted a systematic literature review and three interviews with experts from the industry to collect requirements for a generic point cloud labeling tool. While the literature focused mostly on improvements in labeling time, annotation quality and ease-of-use, the industry experts also emphasized aspects like learnability and the possibility to further adapt and extend the software to individual applications.

*Software Architecture*

Based on the defined user requirements the software was designed in a modular fashion and developed with the flexible Python programming language. This allows the integration of external modules like NumPy and Open3D [6], which excel at point cloud processing, respectively array calculations. Overall, labelCloud is structured according to the MVC design paradigm into the three main compartments model, view, and control. The model captures the representation of the point clouds and labels. labelCloud can import seven different point cloud formats from LiDAR sensors (uncolored) as well as depth cameras (colored). We make use of the Open3D library for most formats and implement a custom loader based on NumPy for binary files (*.bin). Each label is associated with one point cloud and can contain multiple 3D bounding boxes. Each bounding box consists of 10 parameters: 1 for the object class, 3 for the location (x, y, z), 3 for the dimensions (length, width, height), and 3 for the rotations (roll, pitch, yaw). Consequently, users can accurately label objects together with their size and full pose. The created labels can be exported in four different formats including the KITTI [3] format, allowing easy integration with existing frameworks.

The size and structure of point clouds make the navigation and interaction a computationally expensive task. Thus, labelCloud's view makes use of the parallel processing power from the GPU and uses OpenGL for a fluent visualization. The point cloud data is transferred at the beginning of each labeling task and any transformation is realized using the projection matrix. This setup allows the fluent annotation of the often large point cloud files (commonly around 100k points) with standard computer hardware. The point cloud is rotated (left click) and translated (right click) using prevalent mouse commands. Additionally, the user interface provides buttons and text fields for visual user interaction (see Fig. 1).

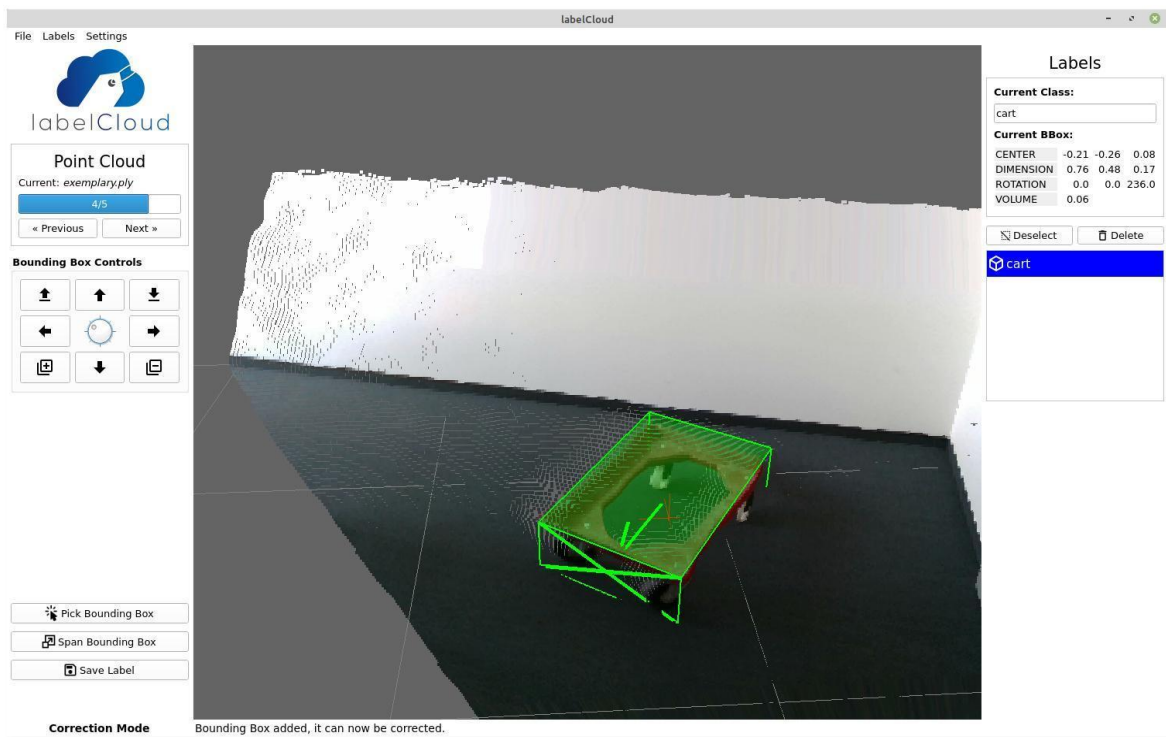

Fig. 1: User interface of labelCloud with a labeled point cloud.

The software behavior is encapsulated in three specialized modules inside the controller compartment. The point cloud manager is responsible for the import, navigation, and manipulation of the point clouds. It keeps track of the progress (share of point clouds labeled) and maps to the underlying file structure. The drawing manager abstracts the different labeling methods. It always has one active labeling mode to which user interactions are forwarded to and is responsible for the live preview of the drawn bounding box. We implement two labeling modes and provide a template method with a standard interface, so that developers can add new custom labeling methods. The label manager exports the annotated labels (bounding box + class) to a file in the chosen format. Through its abstract design, it allows an easy adaptation of existing or extension of additional behavior. With this design, we encourage developers to base more specialized labeling tools on our foundation.

*The Labeling Process*

The labeling process usually consists of three phases: (i) object detection, (ii) bounding box creation, and (iii) parameter correction. Especially in uncolored point clouds, it can take a long time to locate and identify an object. Once that has been done, the user has to enter the object class and create an initial bounding box. While 2D bounding boxes could be spanned using only two clicks, for 3D bounding boxes the object position, size and rotation have to be specified. A labeling method can either solve all steps at the same time or only provide an initial draft of the bounding box. We implement two labeling methods *picking* and *spanning* as well as several possibilities to subsequently improve the parameters of the created bounding box.

The picking mode is based on the assumptions that object sizes are previously known or do not vary too much. It provides a default bounding box with fixed dimensions, which the user can simply drag and rotate into the point cloud. As point clouds have a third dimension, the default bounding box automatically adjusts its size if objects are further in distance. The z-rotation of the bounding box can be adjusted by scrolling the mouse wheel. A preview gives the user a live preview of how the resulting label would look like. Once the location is specified, all other parameters can be freely adjusted.

With the spanning mode, we tried to lift common 2D labeling methods into the 3D space. Instead of selecting two opposing rectangle corners, the user spans the 3D bounding box using four clicks (cf. Fig. 2 a-d). Once the user has picked two vertices, the selection is supported by locking specific dimensions to first specify the object's depth and finally its height. The locking allows the user to also choose points not belonging to the object as long as they represent the desired depth or height. Our evaluation with test users shows that the spanning mode significantly reduces the necessary user interactions by specifying nine parameters using solely four clicks (see Subsection "Evaluation").

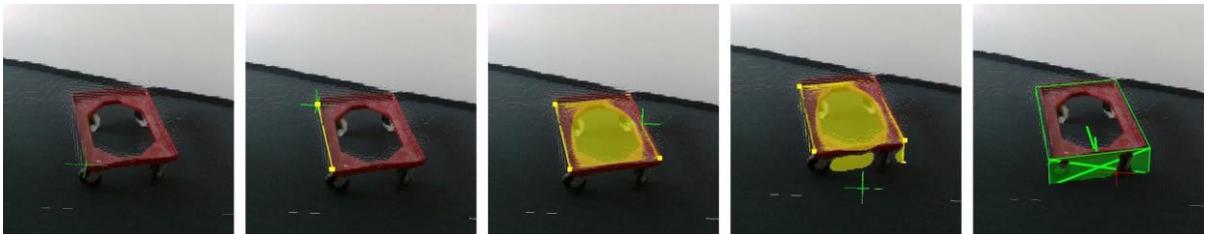

Fig. 2: Task sequence of the spanning mode. Dimensions are locked for the last two points (c + d).

Once an initial bounding box has been created, its parameters can be corrected using a selection of key combinations and visual buttons. Furthermore, labelCloud offers a new user interaction paradigm called "side pulling". As it is very labor-intensive to manually define each object dimension, we allow individual changes in length, width, height using the mouse. The user just has to hover the cursor over the specific bounding box side and can then use the mouse wheel to push or pull the selected side, thus adjusting the perpendicular dimension.

*Point Selection & Depth Estimation*

The labeling interactions inside the viewer require the selection of specific points from the point cloud. However, three-dimensional selection with only two-dimensional visualization (screen) and input devices (mouse) is a difficult problem as mouse clicks only return information about two dimensions (x and y). We overcome this challenge by estimating the third dimension (z) based on cues and assumptions about the user intent. Most software solves this problem using ray casting, i.e. taking the depth of the first object that intersects with a perpendicular ray shot from the click coordinates. This approach fails for point clouds due to occlusion and sparsity, resulting in a volatile selection process. Depth sensors naturally produce sparse point clouds as the scene is recorded from a fixed perspective, but now the user can also take other perspectives. Consequently, there is a high chance the ray will not cast any point and go on into infinity. To tackle these problems, we introduce two assumptions about the user intent and operate on the depth buffer of OpenGL, which holds a depth value for every screen pixel: (i) the user always wants to select a point from the point cloud (i.e., never wants to select infinity), and (ii) the user is more likely to select the closest point to the screen (i.e., on the object surface).

Based on these assumptions we introduce *depth smoothing* and *depth minimization*. Depth smoothing attempts to solve the sparsity problem if the user fails to select a point. This case is detected using a threshold value and leads to the averaging of all depth values in a specified radius around the mouse click. The effect of this method can be compared to the snapping feature in other CAD software (like AutoCAD) and leads to a successful selection, even if the user misses the point.

Depth minimization, on the other hand, is always employed when the user actually clicks on a point. In that case, we use the fact that labeling mostly requires the selection of points on the outer bounds of object surfaces. As the user probably directly faces the object of interest, we assume he or she wants to select the points closest to the screen, thus having a minimal depth value. To reduce surprising effects of this feature, the minimization is done with a smaller radius than the smoothing. Depth minimization is designed to reduce the need for subsequent bounding box corrections.

### *Evaluation*

A first user evaluation of labelCloud showed that the direct labeling approaches lead to higher precision, as measured with an intersection over union (IoU) than the generation of bounding boxes from a point selection (indirect labeling). Test users not familiar with the topic achieved an average IoU of 67% on an indoor test dataset with rotated objects, taking about a minute per point cloud. While both of labelCloud's annotation modes lead to comparable bounding box precision, the spanning mode required significantly less labeling time (-22%) and user interactions (-63%) compared to the picking mode. Also, in a subsequent questionnaire the users described the spanning mode as the more intuitive and performant approach. In a second evaluation, the technical aspect of the software was tested. As the software should represent a suitable solution to quickly create training data for various domains, it must be robust in loading and operating large point clouds. labelCloud's setup takes less than five minutes, as it only requires four selected external libraries, and its source is smaller than 1 MB. It can be downloaded and installed using two command-line instructions on all common operating systems. During performance tests on a standard consumer notebook, it could seamlessly load and handle point clouds with up to five million points, while still rendering with more than 30 FPS. This proves that it is also suitable for domains with extremely large data, like aerial observation.

**Conclusion**

Labeling objects in 3D point clouds is a crucial task to generate training data for ML models in various domains. Existing tools for point cloud labeling are typically designed for limited settings and do not incorporate, amongst others, aspects of no-frills and 3D spanning possibilities. Therefore, we introduced labelCloud which allows for lightweight, convenient 3D point labeling with many different formats without a focus on a specific domain. Our first evaluations show an efficiency increase in comparison to indirect labeling approaches. In future iterations, we plan to integrate transfer learning abilities to recognize patterns of similar objects over time, object tracking as well as additional labeling modes to decrease the average labeling time per point cloud. A demonstration of the tool can be found on YouTube (https://www.youtube.com/watch?v=8GF9n1WeR8A) and the code is available in our GitHub repository (https://github.com/ch-sa/labelCloud).